\documentclass{article}

\usepackage{arxiv}

\usepackage[utf8]{inputenc} 
\usepackage[T1]{fontenc}    
\usepackage{hyperref}       
\usepackage{url}            
\usepackage{booktabs}       
\usepackage{amsfonts}       
\usepackage{nicefrac}       
\usepackage{microtype}      
\usepackage{lipsum}		
\usepackage{graphicx}
\usepackage{natbib}
\usepackage{doi}

\hypersetup{colorlinks,linkcolor=blue,citecolor=blue}
\usepackage{multirow, multicol, booktabs, subfigure}
\usepackage{pifont, dsfont}
\usepackage{amsmath,amssymb}
\usepackage[ruled,linesnumbered]{algorithm2e}
\def\BibTeX{{\rm B\kern-.05em{\sc i\kern-.025em b}\kern-.08em
    T\kern-.1667em\lower.7ex\hbox{E}\kern-.125emX}}
\usepackage{natbib}
\setcitestyle{numbers,square}

\title{Slide-based Graph Collaborative Training for Histopathology Whole Slide Image Analysis}

\date{October 14, 2024}	

\author{Jun Shi\\
	School of Software\\
	Hefei University of Technology\\
	Hefei 230601, China \\
	\And
	Tong Shu \\
	School of Computer Science and Information Engineering\\
	Hefei University of Technology\\
	Hefei 230601, China \\
    \And
	Zhiguo Jiang \\
	Image Processing Center, School of Astronautics\\
	Beihang University\\
	Beijing, 102206, China \\
    \And
	Wei Wang \\
	Department of Pathology, the First Affiliated Hospital of USTC\\
    University of Science and Technology of China\\
	Hefei 230036, China \\
    \And
	Haibo Wu \\
	Department of Pathology, the First Affiliated Hospital of USTC\\
    University of Science and Technology of China\\
	Hefei 230036, China \\
	\texttt{wuhaibo@ustc.edu.cn} \\
    \And
	Yushan Zheng \\
	School of Engineering Medicine\\
    Beijing Advanced Innovation Center on Biomedical Engineering\\
    Beihang University\\
	Beijing 100191, China \\
	\texttt{yszheng@buaa.edu.cn} \\
}

\renewcommand{\shorttitle}{\textit{arXiv} Template}

\hypersetup{
pdftitle={A template for the arxiv style},
pdfsubject={q-bio.NC, q-bio.QM},
pdfauthor={David S.~Hippocampus, Elias D.~Striatum},
pdfkeywords={First keyword, Second keyword, More},
}

\begin{document}
\maketitle

\begin{abstract}
The development of computational pathology lies in the consensus that pathological characteristics of tumors are significant guidance for cancer diagnostics. Most existing research focuses on the inner-contextual information within each WSI yet ignores the possible inter-correlations between slides. As the development of tumors is a continuous process involving a series of histological, morphological, and genetic changes that accumulate over time, the similarities and differences between WSIs across various stages, grades, locations and patients should potentially contribute to the representation of WSIs and deserve to be taken into account in WSI modeling. To verify the advancement of introducing the slide inter-correlations into the representation learning of WSIs, we proposed a generic WSI analysis pipeline SlideGCD that can be adapted to any existing Multiple Instance Learning (MIL) frameworks and improve their performance. With the new paradigm, the prior knowledge of cancer development can participate in the end-to-end workflow, which concurrently initializes and refines the slide representation, as a guide for message passing in the slide-based graph. Extensive comparisons and experiments are conducted to validate the effectiveness and robustness of the proposed pipeline across 4 different tasks, including cancer subtyping, cancer staging, survival prediction, and gene mutation prediction, with 7 representative SOTA WSI analysis frameworks as backbones.
\end{abstract}


\section{Introduction}
{H}{istopathological} characteristics of tumors, including the tendency of tissue invasion, metastasis, growth pattern, etc., have been proven to effectively guide cancer diagnosis and therapies by numerous studies and practices \cite{lu2021data}. Currently, whole slide images (WSIs) have been closely involved in medical practice as an indispensable part of the routine diagnostic process, becoming the gold standard for cancer diagnosis. In recent years, a large amount of research has focused on using artificial intelligence (AI) technology, especially deep learning, on examining WSIs and assist pathologists in effective, accurate, and reproducible pathological analysis and diagnosis, and has achieved significant accomplishments in various fields, e.g. cancer subtyping\cite{zhang2022dtfd, bontempo2023graph}, cancer staging\cite{guo2023higt,shi2023structure}, survival prediction\cite{fan2022cancer,10061470}, gene mutation prediction\cite{zheng2024partial,shi2024masked}, etc.

Considering the special attributes (the giga-pixel resolution and the pyramid structure) that distinguish WSIs from natural scene images, the current WSI analysis framework follows the Multiple Instance Learning (MIL) paradigm which takes patches as the smallest instance of analysis and explores the inner-contextual information of WSI by modeling the correlation between patches. Patch-based WSI analysis methods focus on how to model the relationships between patches more comprehensively and efficiently, and it can be divided into the following four categories: 1) Classical MIL methods \cite{ilse2018attention,lu2021data} treat each patch as an independent instance and generate slide-level representation by aggregating patch-level embeddings via different pooling methods. 2) A series of pseudo-bag based methods \cite{zhang2022dtfd, liu2024pseudo} has been proposed which divide the patches of each WSI into many separated pseudo-bags for solving data scarcity of annotated WSIs. 3) The graph-based methods \cite{li2018graph,chen2021whole,guan2022node,lu2022slidegraph+,shi2023structure, shi2024masked} utilize the patch-based graph where patches are nodes and edges indicate the potential connections between them to simulate the relationships between patches and to represent WSI. 4) The sequence-based methods \cite{shao2021transmil,fillioux2023structured, yang2024mambamil} consider WSI as a sequence of patches and involve various mechanisms or modules, e.g. Transformer or Structured State Space Models, to construct the detailed correlation among patches. With great achievements made by the above studies, the intra-relationships among patches from the same magnification are well-explored, and the interactions across magnifications are arousing more and more attention in recent years \cite{guo2023higt,chan2023histopathology,bontempo2023graph}.

Although the inner-contextual information of WSIs is well-delved by previous research, the inter-correlations between WSIs have not drawn much attention. As most tumors develop through a continuous process involving a series of histological \cite{rivera2014histological}, morphological \cite{simon2016colorectal} and genetic changes \cite{seferbekova2023spatial} that accumulate over time, the similarities and differences between WSIs that across various stages, grades, locations and patients should potentially contribute to the analysis of WSIs and deserve to be taken into account for attaining better slide representations. Some studies \cite{guan2022node,fan2022cancer,shao2023hvtsurv} are aware of the importance of the inter-correlations in patch-slide-patient hierarchy yet stop within the patient level.

\begin{figure}
\centerline{\includegraphics[width=0.8\columnwidth]{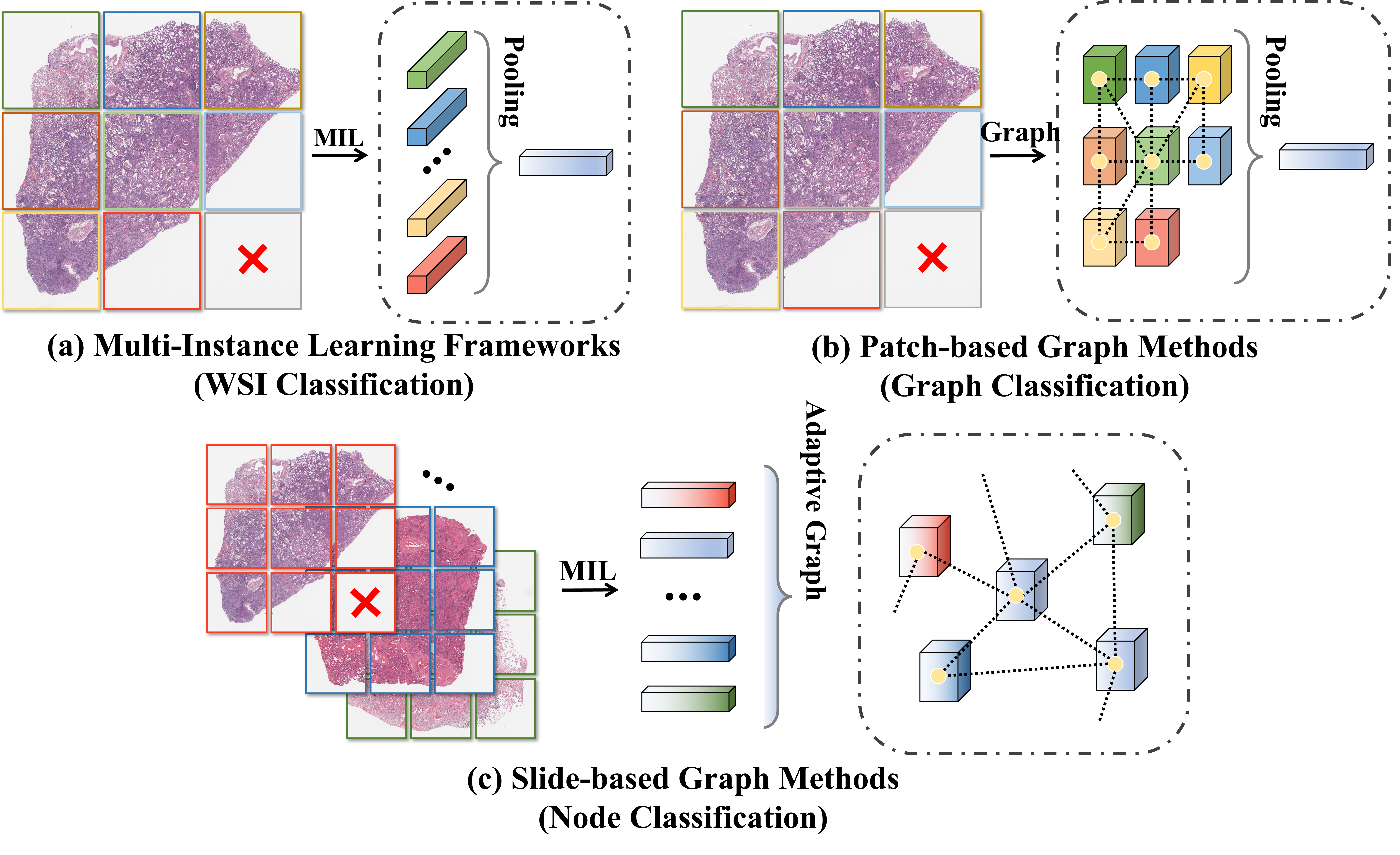}}
\caption{Motivation of our methods. (a) Multi-Instance Learning Frameworks. The main difference between MIL methods lies in the implementations of pooling operations. (b) Patch-based Graph Methods. The mainstream graph-based methods represent WSIs to graphs and transfer the WSI classification problem as graph classification. (c) Slide-based Graph Methods. SlideGCD conceptualizes the WSI classification problem as node classification to explore the inter-correlations between slides explicitly via GNNs.}
\label{Introduction of the slide-based graph}
\end{figure}

In this paper, we explore inter-correlations between WSIs on a larger scope and find an efficient way to unite inter-correlations with the intra-correlations in each WSI. Specifically, we propose the \textbf{S}lide-based \textbf{G}raph \textbf{C}ollaborative training pipeline with knowledge \textbf{D}istillation (SlideGCD) for WSI representation learning that dynamically organizes the slide-level embeddings into a slide-based graph and makes message passing between connected slides via graph neural networks. The intuitive differences between the patch-based graph and the slide-based graph are shown in Fig. \ref{Introduction of the slide-based graph}. More concretely, we take existing MIL methods as the backbone to obtain the initial slide-level embeddings. Then, SlideGCD is used to explore the contextual information implied in the extensive slide-based graph. Finally, the slide-level predictions are obtained by conducting node classification on the slide-based graph.

The main contributions of this paper are summarized below:
\begin{itemize}
\item We propose a new histopathology WSI analysis paradigm that involves prior knowledge of cancer development with coordinatively constructing the slide-based graph and conducting graph message passing during the representation learning. 

\item We devise a rehearsal-based adaptive graph construction strategy to model the slide-level inter-correlations. Besides, a knowledge distillation (KD) based collaborative training for the slide-based hypergraph convolutional network is applied to transfer and enhance the intra-contextual information learned by the MIL network.

\item We conduct extensive comparisons and experiments to validate the effectiveness and generalization of the proposed pipeline across 4 different downstream tasks and 7 representative SOTA WSI analysis frameworks as backbones.
\end{itemize}

\section{Related Works}
\label{Related Works}
This section reviews the approaches related to graph-based WSI analysis and the methods that potentially explore the slide-level inter-correlations.

\subsection{Graph-based WSI Analysis}
Due to its flexibility and interpretability, much attention has been put on the graph structure and graph neural networks. Graph-based methods have shown competitive performance on various WSI analysis tasks. According to different graph structures applied, existing methods can be grouped into the following three categories.

\begin{figure}[!t]
\centerline{\includegraphics[width=1\columnwidth]{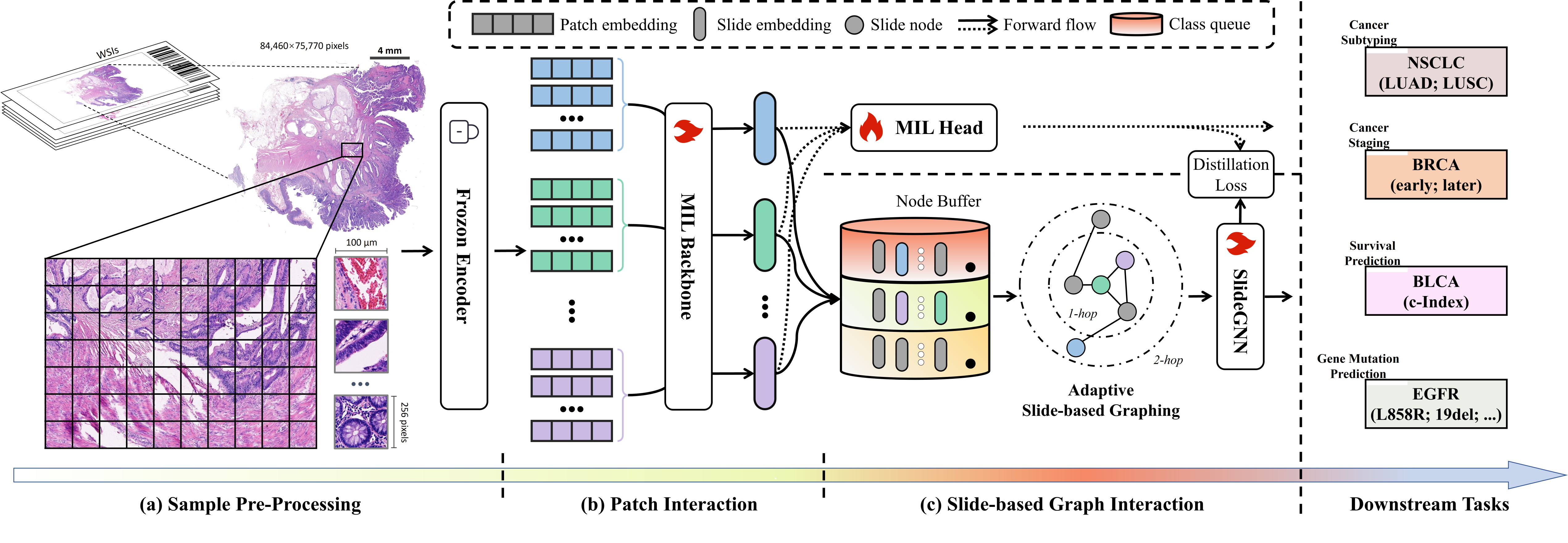}}
\caption{Illustration of the proposed SlideGCD framework. The framework consists of three phases, (a) Sample Pre-Processing. Each WSI is transformed into a sequence of patch embeddings following the universal settings of the MIL paradigm. (b) Patch Interaction. Slide embeddings are generated by the backbone MIL method for each sample. (c) Slide-based Graph Interaction. A slide-based graph is maintained and updated during each mini-batch training (colored indicators denote samples from the current mini-batch), then graph learning and knowledge distillation are conducted to explore slide-level correlations and align both branches.}
\label{Illustration of SlideGCD framework}
\end{figure}

\subsubsection{Methods on Regular graphs}
On account of the analogy that WSI is the graph where its patches are the nodes, graph structures were introduced into WSI analysis at its very early stage. DeepGraphConv \cite{li2018graph} randomly samples 1000+ patches and connects them with a feature similarity threshold to construct a patch-based graph for each WSI. PatchGCN \cite{chen2021whole} builds patch-based graphs via spatial adjacent patches and gains better performance on the survival prediction task. LAMIL \cite{reisenbuchler2022local} and GTP \cite{zheng2022graph} follow the graph construction strategy of PatchGCN yet design different graph transformers to make efficient message passing. NAGCN \cite{guan2022node} introduces the hierarchical global-to-local patch-based graph to represent WSI in both spatial and embedding space. SlideGraph+ \cite{lu2022slidegraph+} uses the biomarker attributes and neural network embeddings to build patch-based graphs and represent the complex organization of cells and the overall tissue micro-architecture. As the idea of multi-scale is progressively involved in WSI analysis, many graph-based methods have also kept up with this trend. SGMF \cite{shi2023structure} constructs the structure-aware hierarchical graph that considers tissue regions and patches from different magnifications in an interactive way. DAS-MIL \cite{bontempo2023graph} creates patch-based graphs on various magnifications and designs a knowledge distillation mechanism to align the representation learned from different graphs.
\subsubsection{Methods on Graph-variant}
Simple graphs consider all nodes equally and their connections can only describe pair-wise relationships. To deal with such drawbacks that deface representation learning in real-world applications, such as WSI representation, the heterogeneous graph and the hypergraph are engaged. HEAT \cite{chan2023histopathology} utilizes a heterogeneous graph with various pre-defined types of nodes to exploit the heterogeneity within WSI and perform WSI classification. Di et al. successively proposed two hypergraph-based WSI analysis frameworks b-HGFN\cite{di2022big} and HGSurvNet\cite{di2022generating}. b-HGFN focuses on efficiently processing the hypergraphs with large-scale vertices, where hyperedges are constructed by adopting the K-nearest neighbor ($k$-NN) strategy on embedding space. HGSurvNet establishes the multi-hypergraph composed of topology-wise sub-hypergraph and phenotype-wise hypergraph for survival prediction with WSIs. MaskHGL \cite{shi2024masked} refines the hypergraph construction strategy with global alignment and designs a mask-reconstruction mechanism for achieving better performance on cancer subtyping and gene mutation prediction.
\subsubsection{Methods on Adaptive graphs}
Apart from the application of the variants of the graph in structure, the adaptive graph where its edge connections could be altered during training is getting attention as well. Hou et al. \cite{hou2022spatial} design a dynamic structure learning module to assist the proposed spatial-hierarchical graph neural network in learning multi-scale information in WSIs. Liu et al. \cite{liu2023graphlsurv} develop a survival-aware structure learning module to construct the adaptive graph for global WSI representation calculation along with the fixed initial graph. WiKG \cite{li2024dynamic} conceptualizes WSIs as a form of knowledge graph structure and dynamically builds the edges via a knowledge-aware attention mechanism during training.

Compared with the previous patch-based graph methods mentioned above, this work exploringly molds the WSIs into a slide-based graph with a rehearsal-based adaptive graph construction module to exploit the slide-level inter-correlations implied in the continuous changes during tumor development.
\subsection{Slide-level Inter-Correlation Exploration}
In clinical practice, each patient probably has multiple WSIs. How to obtain more accurate diagnostic results through multiple slide-level predictions has become a concern for many colleagues. Fan et al. \cite{fan2022cancer} propose an aggregation module that takes slide-level embeddings produced by the front MIL framework to generate patient-level prediction. HVTSurv \cite{shao2023hvtsurv} and P$\&$SrE \cite{li2023patients} choose the Transformer model for interaction between the patient and its belonging slides. The difference is that HVTSurv cascades three transformer blocks for patch-level, WSI-level, and patient-level interaction respectively, yet P$\&$SrE considers that the patient-level embeddings and slide-level embeddings are equal and thus utilizes a single transformer block for their interaction. In addition, some other methods, although not intentionally focusing on the slide inter-correlation, potentially acknowledge the consistency among slides. NAGCN \cite{guan2022node}, HEAT \cite{chan2023histopathology} and MaskHGL \cite{shi2024masked} are aware that there should be a consistency between the constructed graphs in structure or node types. In their graph construction strategies, NAGCN and MaskHGL apply a hierarchical clustering method based on all the patches no matter which WSI it comes from to align the graph structure across slides. HEAT employs a pre-trained network to classify the patches into pre-defined node types thus achieving node-level consistency.

From the perspective of slide-level inter-correlation exploration, the above methods are not comprehensive enough, as they either involve slide-level interactions limited within the single patient or do not model the slide-level inter-correlations in an explainable way. The proposed SlideGCD lifts the patient-level restriction and explicitly constructs a slide-based graph to explore the contextual information.

\section{Methodology}
\label{Methodology}

In this section, we describe our proposed framework which consists of three phases: sample pre-processing, patch interaction, and slide-based graph interaction, as illustrated in Fig. \ref{Illustration of SlideGCD framework}. In the first phase, each WSI sample is processed into a sequence of patch embeddings with a pre-trained patch encoder. For the patch interaction phase, we engage the existing MIL method as the backbone to generate slide-level embeddings. In the slide-based graph interaction phase, a rehearsal-based adaptive graph construction strategy is exploited to build and maintain a slide-based graph during training. Then, the SlideGNN is deployed to explore the slide inter-correlation based on the slide-based graph and refine the slide embeddings for solving downstream tasks more precisely. Additionally, an online distillation is designed between the MIL head and the SlideGNN to solve the problem of knowledge misalignment.

\subsection{Sample Pre-Processing}
Assuming there is a dataset with $N$ WSIs denoted as $\mathcal{D} = \{(\mathbf{B}_i, y_i)\}_{i=1}^N$. Each WSI $\mathbf{B}_i = \{I_{i, j}\}_{j=1}^{M_i}$ is annotated with a label $y_i \in \{0, ..., C - 1\}$, where $I_{i, j}$ is tiled patch without patch-level label, $M_i$ is the number of patches and $C$ represents the number of categories. Then, there is a pre-trained patch encoder $f(\cdot)$, where we used PLIP \cite{huang2023visual}, to transform the patch $I_k$ into patch embeddings $\textbf{x}_k \in \mathbb{R}^{512}$.

\subsection{Patch Interaction}
After obtaining the patch embeddings, an aggregator network is needed to exploit the patch correlations and generate slide embeddings. We leave this job to the existing MIL network which will be called Backbone in our following statement. In an ideal implementation, the Backbone can be any MIL network with any architecture as long as it can produce fixed $D_s$ dimensional slide-level embeddings $\mathbf{s}_i \in \mathbb{R}^{D_s}$. The MIL Head is the relevant sub-network of the Backbone for making predictions $\hat{\mathbf{y}}^{MIL}$ for downstream tasks. Note that the slide embeddings are quite unstable during the first few training epochs. Such instability might defect the subsequent graph learning. Therefore, we set a few warmup epochs at the beginning of the training for backbone pre-convergence, in which only the MIL network is involved in forward and backward. 

We employed several representative MIL methods as the Backbone and conducted different downstream tasks for the proposed SlideGCD, more analysis can be found in Section \ref{Experiments}.

\subsection{Slide-based Graph Interaction}
With the slide embeddings, the requirements for slide-based graph interaction have been fulfilled. In this section, we describe the main contributions including the rehearsal-based adaptive graph construction strategy, the details of the designed SlideGNN, and the collaborative training with knowledge distillation.

\subsubsection{Rehearsal-based Adaptive Graph Construction}
With the inspiration of the idea of the Memory Bank \cite{he2020momentum,li2022lesion} and the rehearsal-based continual learning \cite{de2021continual,huang2023conslide}, a Class-Aware Node Buffer is designed to store the previous slide embeddings which will participate in the slide-based graph construction and will be replayed in graph learning. Specifically, the Class-Aware Node Buffer defines a storage space $\mathbf{Q}^T = [\mathbf{Q}_{11}^T, \mathbf{Q}_{12}^T, ..., \mathbf{Q}_{1C}^T]$, where $\mathbf{Q} \in  \mathbb{R}^{L \times D_s}$ represents that it is able to deposit $L$ slide embeddings and partitioned matrix $\mathbf{Q}_{1c} \in \mathbb{R}^{\frac{L}{C} \times D_s}$ indicates the sub-queue that is responsible for storing slide embeddings with $c$-th category. Each slide embedding stored in the buffer or the current mini-batch will correspond to a node in the subsequent slide-based graph and will be considered as its initial node embedding. 

\begin{figure}
\centerline{\includegraphics[width=0.7\columnwidth]{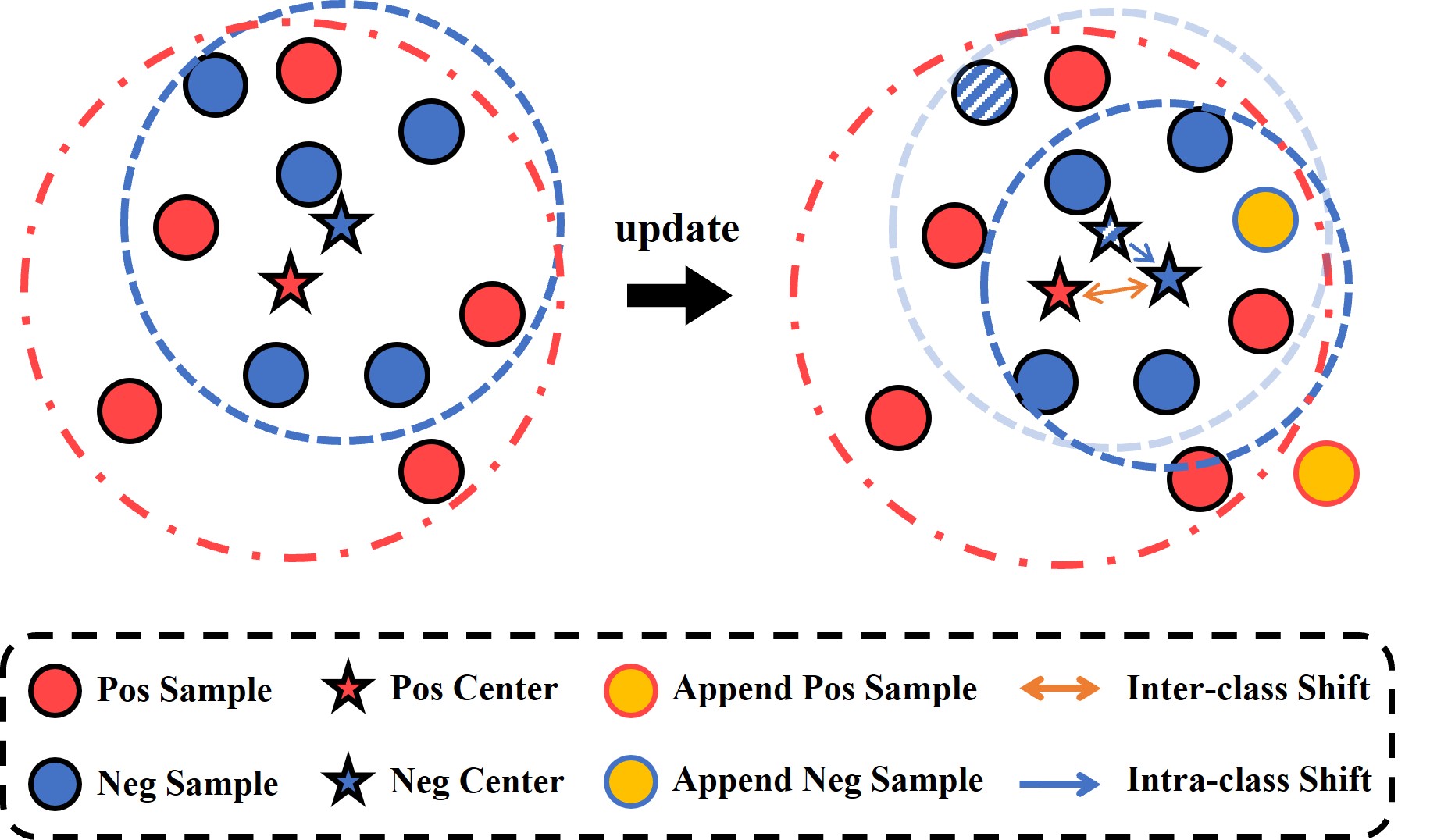}}
\caption{Illustration of the update of the Node Buffer in embedded space. Left: the initial state of the Node Buffer after the warmup, where the centers of each category are close. Right: the updated state after a mini-batch with 2 samples. The appended positive sample is too far from the center even compared with the farthest stored node thus it won't be updated into the buffer since we consider it as an amplified disturbance from the former network update. The appended negative sample is close enough to the center to replace the farthest negative node marked by stripes. At last, the outcome of the update is that the negative center shifts away from the positive center as we expected.}
\label{Illustration of BufferUpdate}
\end{figure}

In addition, the rehearsal buffer will be updated during each mini-batch of training.
During the warmup stage, we directly apply the First-In-First-Out (FIFO) strategy to update the node buffer, in which the newest mini-batch of slide embeddings will be randomly pushed into the node buffer and the outdated slide embeddings at the end of the buffer will be popped out simultaneously. In the formal training stage, we first calculate the centers of each sub-queue in the embedded space, and then each slide embedding in the current mini-batch is going to replace the farthest stored sample in the corresponding sub-queue as long as it is closer. As updates continue, ultimately the sub-queues will be separated from each other. The following loss function is applied to ensure that during formal buffer updating. 

\begin{align}
    L_{update} = - \sum_{\mathbf{u} \in \mathbf{U}} \log\frac{\exp(\mathbf{u} \cdot \textbf{q}_+ / \tau)}{\sum_{i=0}^{C}\exp(\mathbf{u} \cdot \textbf{q}_i / \tau)}  +  \notag
    \\ \sum_{\mathbf{u} \in \mathbf{U}} \sum_{i=0}^C \mathds{1}_{\textbf{q}_i\neq\textbf{q}_+}(\frac{\mathbf{u} \cdot \textbf{q}_i}{\Vert \mathbf{u} \Vert \Vert \textbf{q}_i \Vert}),
\end{align}
where $\mathbf{U}$ is the enqueued slide embeddings, $\textbf{q}_i$ indicates the center of $i$-th sub-queue, the $\textbf{q}_+$ represents the center which corresponds to the category of $\mathbf{u}$ and the $\tau$ is the temperature coefficient. $\mathds{1}_{\textbf{q}_i\neq\textbf{q}_+}(\cdot)$ means that this term is not 0 only if $\textbf{q}_i\neq\textbf{q}_+$. An illustration of the update is shown in Fig. \ref{Illustration of BufferUpdate}.

After getting the preliminary separable nodes, we conduct the adaptive graph generation (AGG) strategy to infer the inter-dependencies from the embedding space for connecting these nodes with hyperedges. Our AGG strategy consists of a linear layer that transforms the slide embeddings into an intermediate hidden space and a $k$-NN clustering operator that will be performed on the intermediate embeddings to connect each node with its $k$ nearest neighbors with a hyperedge. Eventually, the slide embeddings in the current mini-batch and the slide embeddings retrieved from the node buffer are formulated as a hypergraph $\mathcal{G}$. The adaptive graph generation process can be formulated as:

\begin{equation}
    \mathcal{G} = (\mathbf{X}^{(0)}, \mathcal{E}),\  \mathcal{E} = {\rm KNN}(\mathbf{P}, k), \ \mathbf{P} = {\rm Linear}(\mathbf{X}^{(0)})
\end{equation}
where $\textbf{X}^{(0)} \in \mathbb{R}^{(L+B)\times D_s}$ is the node embeddings sequence that consists of the data in the buffer with a length of $L$ and the current mini-batch data with a length of $B$ (batch size) and $\mathcal{E}$ is the set of hyperedges that describes the connections between nodes.

\subsubsection{Graph Learning via SlideGNN}
With the slide-based graph representation, the SlideGNN composed of two hypergraph convolutional layers \cite{bai2021hypergraph} and a Centering-Attention module is applied to explore the implied information, as:
\begin{equation}
    \mathbf{X}^{(i+1)} = {\rm LeakyReLU}({\rm HGC}(\mathbf{X}^{(i)}, \mathcal{E})),
\end{equation}
\begin{equation}
    \mathbf{H} = {\rm Concat}(\mathbf{X}^{(0)}, \mathbf{X}^{(1)}, \mathbf{X}^{(2)}), \mathbf{H} \in \mathbb{R}^{L \times 3D_S},
\end{equation}
where ${\rm HGC}(\cdot)$ denotes the hypergraph convolution and $\mathbf{X}^{(i)}$ contains the information accumulated from the node itself to its $i$-hop neighbors.

Given that the adaptive graphs can involve graph heterophily that nodes with different categories are connected, we designed the Centering-Attention module to alleviate such heterophily by rebalancing the participation of $k$-hop information. The Centering-Attention module is implemented with channel-wise attention and $Centering$ operation \cite{Caron_2021_ICCV} which prevent the attention score from always being positive even when facing defective partial information. The computations can be formulated as:
\begin{equation}
    \mathbf{H}' = \mathbf{H} \cdot Centering(\mathbf{a}) = \mathbf{H} \cdot (\mathbf{a} - {\rm Mean}(\mathbf{a})),
\end{equation}
\begin{equation}
    \mathbf{a} = {\rm Sigmoid}({\rm ReLU}(\mathbf{H}^\mathrm{T}\mathbf{W}_0)\mathbf{W}_1),
\end{equation}
where $\mathbf{W}_0, \mathbf{W}_1 \in \mathbb{R}^{L \times L}$ are the learnable weights, $\mathbf{a} \in \mathbb{R}^{3D_S}$ is the attention score before $Centering$. Finally, an MLP classifier $Cls_{Graph}$ with one linear layer is used to make final predictions $\hat{\mathbf{y}}^G$ for current mini-batch:
\begin{equation}
    \hat{\mathbf{y}}^G = {\rm Softmax}({\rm Linear}(\mathbf{H}')),
\end{equation}

\subsubsection{Collaborative Training with Knowledge Distillation}
\label{Training workflow}
With the rehearsal-based adaptive graph and the SlideGNN, this framework is capable of exploiting the slide inter-correlations from the given WSI dataset. However, there is no interaction between the graph branches and the MIL head (Head$_{\rm MIL}$), thus the well-learned slide intrinsic knowledge implied in the Head$_{\rm MIL}$ may be neglected. To associate it with the slide inter-correlations and constrain both branches, we involved the knowledge distillation to transfer the knowledge learned by Head$_{\rm MIL}$ to the SlideGNN.

We treat the Head$_{\rm MIL}$ and the SlideGNN as the teacher and student model separately, letting SlideGNN draw on the beneficial information learned by Head$_{\rm MIL}$. Specifically, a response-based knowledge distillation loss, JS divergence loss, is adopted as:
\begin{equation}
    L_{KD} = \sum_i^Cp^G_i\log(\frac{2p^G_i}{p^G_i + p^{MIL}_i}) + \sum_i^Cp^{MIL}_i\log(\frac{2p^{MIL}_i}{p^G_i + p^{MIL}_i}), \notag
\end{equation}
\begin{equation}
    p^G_i = \frac{exp(\hat{y}^G_i / t)}{\sum_jexp(\hat{y}^G_j / t)},\quad p^{MIL}_i = \frac{exp(\hat{y}^{MIL}_i / t)}{\sum_jexp(\hat{y}^{MIL}_j / t)}
\end{equation}
where $t$ is the temperature coefficient.

Then, the final loss of SlideGCD can be written as below, $L_{CE}(\cdot)$ represents the Cross-Entropy loss function, and $\beta$ is the weight contributed by the buffer update:
\begin{equation}
    L = L_{CE}(\hat{\mathbf{y}}^{MIL}, y) + L_{CE}(\hat{\mathbf{y}}^G, y) + L_{KD} + \beta L_{update}.
\end{equation}

\subsection{Inference}
At the inference stage, all parameters and the Class-Aware Node Buffer are frozen. When a WSI is inputted, 1) its initial embedding will be made with the backbone, 2) the initial slide embedding will be inserted into the slide-based graph with the same rehearsal-based adaptive graph construction strategy, 3) the SlideGNN will make message passing to refine its embedding for final prediction.

\begin{table}
\centering
\caption{Comparisons of the baselines and corresponding SlideGCD collaboration version on cancer subtyping. $\dag$: When reproducing HiGT, we applied the default setting on multi-scale from its original paper that only considered the thumbnail, 5$\times$, and 10$\times$ magnifications. That is the reason why the performance of HiGT gaps with other methods.}
\begin{tabular}{lc|ccc|ccc}
\toprule
\multirow{2}*{Method} & \multirow{2}{*}{SlideGCD} & \multicolumn{3}{c}{TCGA-BRCA}& \multicolumn{3}{c}{TCGA-NSCLC} \\ 
& & ACC(\%) & AUC(\%) & F1(\%) & ACC(\%) & AUC(\%) & F1(\%) \\ \midrule

\multirow{3}{*}{ABMIL \cite{ilse2018attention}} 
& \ding{55} & 88.97$\pm$0.85 & 89.87$\pm$0.89 & 81.61$\pm$2.27
& 86.96$\pm$1.16 & 95.14$\pm$0.51 & 86.89$\pm$1.20\\

& \ding{52} & 89.77$\pm$1.04 & 91.33$\pm$1.52 & 83.87$\pm$1.83
& \textbf{91.04$\pm$2.42} & \underline{97.10$\pm$1.49} & \textbf{91.02$\pm$2.45}\\

& $\Delta$ & \textcolor[RGB]{0, 200, 0}{+0.80} & \textcolor[RGB]{0, 200, 0}{+1.46} & \textcolor[RGB]{0, 200, 0}{+2.26}
& \textcolor[RGB]{0, 200, 0}{+4.08} & \textcolor[RGB]{0, 200, 0}{+1.96} & \textcolor[RGB]{0, 200, 0}{+4.13}\\ \midrule

\multirow{3}{*}{PatchGCN \cite{chen2021whole}} 
& \ding{55} & 84.80$\pm$1.77 & 87.18$\pm$1.55 & 75.11$\pm$4.57
& 86.62$\pm$2.38 & 94.81$\pm$1.82 & 86.59$\pm$2.43\\

& \ding{52} & 86.00$\pm$0.87 & 89.79$\pm$1.30 & 76.49$\pm$0.78
& 89.63$\pm$1.68 & \textbf{97.62$\pm$0.75} & 89.60$\pm$1.69\\

& $\Delta$ & \textcolor[RGB]{0, 200, 0}{+1.20} & \textcolor[RGB]{0, 200, 0}{+2.61} & \textcolor[RGB]{0, 200, 0}{+1.38}
& \textcolor[RGB]{0, 200, 0}{+3.01} & \textcolor[RGB]{0, 200, 0}{+2.81} & \textcolor[RGB]{0, 200, 0}{+3.01}\\ \midrule

\multirow{3}{*}{TransMIL \cite{shao2021transmil}} 
& \ding{55} & 88.17$\pm$1.00 & 90.99$\pm$0.91 & 82.09$\pm$1.75
& 85.82$\pm$1.67 & 94.82$\pm$1.17 & 85.77$\pm$1.67\\

& \ding{52} & \underline{90.70$\pm$1.73} & \underline{92.82$\pm$2.00} & \underline{85.91$\pm$2.62}
& 90.70$\pm$3.00 & 96.53$\pm$1.22 & 90.68$\pm$3.03\\

& $\Delta$ & \textcolor[RGB]{0, 200, 0}{+2.53} & \textcolor[RGB]{0, 200, 0}{+1.83} & \textcolor[RGB]{0, 200, 0}{+3.82}
& \textcolor[RGB]{0, 200, 0}{+4.85} & \textcolor[RGB]{0, 200, 0}{+1.71} & \textcolor[RGB]{0, 200, 0}{+4.91}\\ \midrule

\multirow{3}{*}{DTFDMIL \cite{zhang2022dtfd}} 
& \ding{55} & 89.30$\pm$0.44 & 90.08$\pm$0.86 & 83.17$\pm$1.43 
& 86.42$\pm$1.07 & 95.59$\pm$0.66 & 86.40$\pm$1.06 \\

& \ding{52} & \textbf{91.50$\pm$0.50} & \textbf{92.83$\pm$0.93} & \textbf{86.52$\pm$0.78}
& \underline{90.84$\pm$1.83} & 96.31$\pm$1.84 & \underline{90.82$\pm$1.84} \\

& $\Delta$ & \textcolor[RGB]{0, 200, 0}{+2.20} & \textcolor[RGB]{0, 200, 0}{+2.75} & \textcolor[RGB]{0, 200, 0}{+3.35}
& \textcolor[RGB]{0, 200, 0}{+4.42} & \textcolor[RGB]{0, 200, 0}{+0.72} & \textcolor[RGB]{0, 200, 0}{+4.42}\\ \midrule

\multirow{3}{*}{HiGT \dag \cite{guo2023higt}} 
& \ding{55} & 86.09$\pm$1.13 & 86.20$\pm$0.59 & 77.93$\pm$1.14
& 85.42$\pm$1.96 & 93.93$\pm$0.42 & 85.33$\pm$2.06 \\

& \ding{52} & 88.43$\pm$0.40 & 87.93$\pm$2.07 & 81.47$\pm$0.34
& 87.32$\pm$1.53 & 94.65$\pm$0.58 & 87.28$\pm$1.56\\

& $\Delta$ & \textcolor[RGB]{0, 200, 0}{+2.34} & \textcolor[RGB]{0, 200, 0}{+1.73} & \textcolor[RGB]{0, 200, 0}{+3.54}
& \textcolor[RGB]{0, 200, 0}{+1.90} & \textcolor[RGB]{0, 200, 0}{+0.72} & \textcolor[RGB]{0, 200, 0}{+1.95}\\ \midrule

\multirow{3}{*}{S4MIL \cite{fillioux2023structured}} 
& \ding{55} & 87.84±0.50 & 89.29±1.71 & 81.18±1.00
& 88.03±0.86 & 95.05±0.70 & 88.01±0.87 \\

& \ding{52} & 89.30±1.62 & 90.12±2.34 & 83.35±1.79
& 89.50±1.88 & 96.04±1.42 & 89.48±1.88 \\

& $\Delta$ & \textcolor[RGB]{0, 200, 0}{+1.46} & \textcolor[RGB]{0, 200, 0}{+0.83} & \textcolor[RGB]{0, 200, 0}{+2.17}
& \textcolor[RGB]{0, 200, 0}{+1.47} & \textcolor[RGB]{0, 200, 0}{+0.99} & \textcolor[RGB]{0, 200, 0}{+1.47}\\
\bottomrule
\end{tabular}
\label{Performance on cancer subtyping}
\end{table}

\section{Experiments}
\label{Experiments}
\subsection{Datasets}
We conducted extensive experiments on various downstream tasks to verify the effectiveness of the proposed pipeline, including cancer subtyping, cancer staging, survival prediction, and gene mutation prediction.
\subsubsection{Cancer Subtyping} Two publicly available WSI datasets are used to evaluate our proposed pipeline in the downstream task of cancer subtyping, including the TCGA-BRCA and the TCGA-NSCLC released by The Cancer Genome Atlas (TCGA) project \cite{kandoth2013mutational}\footnote{https://portgdc.cancer.gov/}. Speciﬁcally, TCGA-BRCA contains 998 diagnostic digital slides of two breast cancer subtypes, made up of 794 WSIs of invasive ductal carcinoma (IDC) and 204 WSIs of invasive lobular carcinoma (ILC). TCGA-NSCLC is a collection of two subtype projects for lung cancer, i.e. lung squamous cell carcinoma (LUSC) and lung adenocarcinoma (LUAD), for a total of 995 diagnostic WSIs, including 496 WSIs of LUSC and 499 WSIs of LUAD.
\subsubsection{Cancer Staging} In the cancer staging task, we used the same WSIs with another set of labels for staging from the two public datasets mentioned above, TCGA-BRCA and TCGA-NSCLC. Concretely, we excluded samples without grading labels and categorized the remaining WSIs as early-stage (Stage I\&II) and late-stage (Stage III\&IV). In TCGA-BRCA, there are 713 WSIs of early-stage and 232 WSIs of late-stage. In TCGA-NSCLC, there are 726 WSIs of early-stage and 180 WSIs of late-stage. Note that the staging task is normally harder than subtyping because it relies more on subtle cellular morphological features and the natural class imbalance of its datasets.
\subsubsection{Survival Prediction} As to the survival prediction task, we evaluated SlideGCD with the TCGA-BLCA (437 WSIs) cohort following the setting of previous studies \cite{chen2021whole}. 
\subsubsection{Gene Mutation Prediction} An in-house clinical dataset USTC-EGFR\footnote{The study was approved by the Medical Research Ethics Committee of the First Affiliated Hospital of the University of Science and Technology of China (Anhui Provincial Hospital) under the protocol No.2022-RE-454.} is used to evaluate whether the slide inter-correlation can benefit the efficiency of gene mutation prediction via H\&E histopathology WSIs. USTC-EGFR contains a total of 754 WSIs of lung histopathology from five categories of samples, including 165 WSIs of negative (Neg), 118 WSIs with a missense mutation in exon 21 (L858R), 184 WSIs with in-frame deletions in exon 19 (19del), 146 WSIs of wild type (Wild) and 141 WSIs with other driver gene mutations (Others). All labels have been confirmed by pathologists.

\subsection{Implementation Details}
Before training, the 256$\times$256 patches within tissue regions were split under 20$\times$ lenses. During training, the Adam optimizer with CosineAnnealing learning rate scheduler was employed. We directly adopted the default setting of hyper-parameters to the baseline from its public repository and remained fixed when applying SlideGCD for the fairness of comparison except the learning rate will be reset to 1e-4 after the warmup for SlideGCD. All methods are implemented in Python with the \textit{PyTorch} 1.8 and \textit{PyTorch Geometric (PyG)} libraries. We ran the experiments on a computer with an NVIDIA RTX 3090 GPU. 

All experiments are performed using the same five-fold cross-validation splits. The average accuracy (ACC), macro-average F1 score (F1), and macro-average area under the receiver operating characteristic curve (AUC) are calculated for evaluating the classification performance, i.e. cancer subtyping, cancer staging and gene mutation prediction tasks. The concordance index (C-Index) is calculated to evaluate the performance of the survival prediction task.

\subsection{Results and Analysis}
We evaluate the effectiveness of the proposed SlideGCD by comparing its improvement on various state-of-the-art WSI analysis approaches and different downstream tasks. For the cancer subtyping and the cancer staging tasks, we select 6 influential previous SOTA methods as baselines: \textit{i) \textbf{ABMIL}} \cite{ilse2018attention}, \textit{ii) \textbf{PatchGCN}} \cite{chen2021whole}, \textit{iii) \textbf{TransMIL}} \cite{shao2021transmil}, \textit{iv) \textbf{DTFDMIL}} \cite{zhang2022dtfd}, \textit{v) \textbf{HiGT}} \cite{guo2023higt}, \textit{vi) \textbf{S4MIL}} \cite{fillioux2023structured}. For the survival prediction tasks, we choose 2 classic methods, \textit{i) \textbf{ABMIL-Surv}} and \textit{ii) \textbf{PatchGCN}}, that performed well in the reported results in \cite{chen2021whole} for this task. For the gene mutation prediction, we choose \textit{\textbf{DTFDMIL}} as one of the two selected baselines since it gains the best overall performance on our subtyping and staging experiments. The other one is \textit{\textbf{MaskHGL}} \cite{shi2024masked} for it is one of the latest WSI analysis methods that pay special attention to the fine-grained gene mutation prediction task. The experimental results for each downstream task are presented respectively in Tables \ref{Performance on cancer subtyping}-\ref{Performance on gene mutation prediction}.

\textbf{Overall}, the proposed SlideGCD is capable of substantially improving the accuracy of baseline models in most application scenarios. Especially for the pseudo-bag based method DTFDMIL, the SlideGCD significantly improved its performance on 3 different WSI classification tasks, including but not limited to achieving the best performance on TCGA-BRCA (Subtyping) with 91.50\% ACC, 92.83\% AUC and 86.52\% F1, and achieving over 4\% improvement of ACC \& F1 on TCGA-NSCLC (Subtyping), and 9.77\% of AUC gain on TCGA-NSCLC (Staging).

\begin{table}
\centering
\caption{Comparsions on cancer staging. The ACCs are not reported because the serious class imbalance on cancer staging makes the ACC of all methods approach the proportion of the class with more samples in the test set, thereby making the accuracy less referenceable.}
\begin{tabular}{lc|cc|cc}
\toprule
\multirow{2}*{Method} & \multirow{2}{*}{SlideGCD} & \multicolumn{2}{c}{TCGA-BRCA}& \multicolumn{2}{c}{TCGA-NSCLC} \\ 
& & AUC(\%) & F1(\%) & AUC(\%) & F1(\%) \\ \midrule

\multirow{3}{*}{ABMIL \cite{ilse2018attention}} 
& \ding{55} & 58.75±1.84 & 46.00±3.15
& 63.98±1.96 & 58.24±3.56\\

& \ding{52} & 60.79±3.18 & 48.24±6.36
& 67.57±2.02 & 58.03±6.48\\

& $\Delta$ & \textcolor[RGB]{0, 200, 0}{+2.04} & \textcolor[RGB]{0, 200, 0}{+2.24}
& \textcolor[RGB]{0, 200, 0}{+3.59} & \textcolor[RGB]{100, 100, 100}{-0.21}\\ \midrule

\multirow{3}{*}{PatchGCN \cite{chen2021whole}} 
& \ding{55} & 57.96±1.80 & 49.28±4.61
& 59.66±1.63 & 53.46±5.66\\

& \ding{52} & \textbf{61.17±3.38} & 51.54±1.99
& 65.85±5.17 & 53.36±5.89\\

& $\Delta$ & \textcolor[RGB]{0, 200, 0}{+3.21} & \textcolor[RGB]{0, 200, 0}{+2.26}
& \textcolor[RGB]{0, 200, 0}{+6.19} & \textcolor[RGB]{100, 100, 100}{-0.10}\\ \midrule

\multirow{3}{*}{TransMIL \cite{shao2021transmil}} 
& \ding{55} & 58.25±3.11 & 47.38±4.26
& 61.91±1.51 & 54.64±3.34\\

& \ding{52} & 59.79±3.35 & 51.93±3.70
& 63.47±2.65 & 56.97±4.02\\

& $\Delta$ & \textcolor[RGB]{0, 200, 0}{+1.54} & \textcolor[RGB]{0, 200, 0}{+4.55}
& \textcolor[RGB]{0, 200, 0}{+1.56} & \textcolor[RGB]{0, 200, 0}{+2.33}\\ \midrule

\multirow{3}{*}{DTFDMIL \cite{zhang2022dtfd}} 
& \ding{55} & 58.14±2.52 & 53.72±1.37
& 59.73±1.74 & 54.74±2.72\\

& \ding{52} & 60.23±1.78 & \underline{54.25±2.86}
& \textbf{69.50±3.48} & \underline{58.64±6.23}\\

& $\Delta$ & \textcolor[RGB]{0, 200, 0}{+2.09} & \textcolor[RGB]{0, 200, 0}{+0.53}
& \textcolor[RGB]{0, 200, 0}{+9.77} & \textcolor[RGB]{0, 200, 0}{+3.90}\\ \midrule

\multirow{3}{*}{HiGT \cite{guo2023higt}} 
& \ding{55} & 56.23±1.58 & 50.86±4.44
& 58.77±2.65 & 50.65±2.25\\

& \ding{52} & 56.41±2.85 & 49.72±5.63
& 59.44±3.60 & 50.09±5.78\\

& $\Delta$ & \textcolor[RGB]{100, 100, 100}{+0.18} & \textcolor[RGB]{200, 0, 0}{-1.14}
& \textcolor[RGB]{0, 200, 0}{+0.67} & \textcolor[RGB]{200, 0, 0}{-0.56}\\ \midrule

\multirow{3}{*}{S4MIL \cite{fillioux2023structured}} 
& \ding{55} & 58.55±5.04 & 52.37±2.86
& 62.29±3.25 & 54.65±5.28\\

& \ding{52} & \underline{61.13±2.20} & \textbf{54.69±2.63}
& \underline{67.54±2.35} & \textbf{59.55±5.21}\\

& $\Delta$ & \textcolor[RGB]{0, 200, 0}{+2.58} & \textcolor[RGB]{0, 200, 0}{+2.32}
& \textcolor[RGB]{0, 200, 0}{+5.25} & \textcolor[RGB]{0, 200, 0}{+4.90}\\
\bottomrule
\end{tabular}
\label{Performance on cancer staging}
\end{table}

Each dataset we applied has its characteristics. For example, the dataset (TCGA-NSCLC) used in cancer subtyping corresponds to the most common binary classification without class imbalance. The cancer staging reveals the imbalanced binary classification and the fine-grained gene mutation prediction represents the multi-class classification problem. Moreover, the regression problem is validated by the survival prediction task. All these multi-task experiments have proven the robustness and generalization of the proposed SlideGCD.

\begin{table}
\centering
\caption{Comparisons on survival prediction.}
\begin{tabular}{lc|c}
\toprule
\multirow{2}*{Method} & \multirow{2}{*}{SlideGCD} & TCGA-BLCA \\ 
& & C-Index \\ \midrule

\multirow{3}{*}{ABMIL-Surv} 
& \ding{55} & 52.72±4.55\\

& \ding{52} & 54.29±6.52\\

& $\Delta$ & \textcolor[RGB]{0, 200, 0}{+1.57}\\ \midrule

\multirow{3}{*}{PatchGCN \cite{chen2021whole}} 
& \ding{55} & 55.63±3.91\\

& \ding{52} & 57.45±4.00\\

& $\Delta$ & \textcolor[RGB]{0, 200, 0}{+1.82}\\
\bottomrule
\end{tabular}
\label{Performance on survival prediction}
\end{table}

One exception appears on HiGT, the multi-scale method that makes further consideration in hierarchical interaction across different magnifications via self-attention variant, in the cancer grading task. A foreseeable reason is the implicit contextual information at low magnification (e.g. thumbnails, 5$\times$, and 10$\times$) makes it difficult to achieve accurate cancer grading. Additionally, the unaligned patch clustering in HiGT, which is performed for each WSI separately, disturbs the connection measurement between WSIs and increases the heterogeneity of the slide-based graph. One another unideal situation occurred with MaskHGL on gene mutation prediction where the improvements were not significant enough as well. The problem may be with the Masked Hypergraph ReConstruction (MHRC) module that is responsible for the mask reconstruction in MaskHGL. The learnable mask token in MHRC bridges the different slide representations since it participates in every round of training that involves the mask operation, thus the inter-correlation may have been learned in MaskHGL.

We are also aware that the improvement in the TCGA-BRCA cohort is smaller compared with TCGA-NSCLC in both cancer subtyping and staging settings. Considering the varying difficulty levels of each dataset, the TCGA-BRCA cohort is harder than TCGA-NSCLC. It might suggest that the benefits brought by SlideGCD decline as the difficulty increases. The deeper reason is the separability of node embeddings is affected by the dataset difficulty and that narrows the potential of slide-based graph learning guided by the connection reflecting similar pathological patterns. However, even so, our SlideGCD can still achieve universal improvement in more complex downstream tasks, e.g. survival prediction and fine-grained gene mutation prediction.

\begin{table}
\centering
\caption{Comparisons on gene mutation prediction.}
\begin{tabular}{lc|ccc}
\toprule
\multirow{2}*{Method} & \multirow{2}{*}{SlideGCD} & \multicolumn{3}{c}{USTC-EGFR} \\ 
& & ACC(\%) & AUC(\%) & F1(\%) \\ \midrule

\multirow{3}{*}{DTFDMIL \cite{zhang2022dtfd}} 
& \ding{55} & 57.49±2.13 & 85.47±0.32 & 55.80±1.52 \\

& \ding{52} & 59.91±2.80 & 85.73±1.43 & 59.53±2.91 \\

& $\Delta$ & \textcolor[RGB]{0, 200, 0}{+2.42} & \textcolor[RGB]{0, 200, 0}{+0.26} & \textcolor[RGB]{0, 200, 0}{+3.73}\\ \midrule

\multirow{3}{*}{MaskHGL \cite{shi2024masked}} 
& \ding{55} & 61.52±2.86 & 86.82±0.99 & 60.22±3.58 \\

& \ding{52} & 62.24±3.20 & 87.51±1.31 & 61.56±3.76 \\

& $\Delta$ & \textcolor[RGB]{0, 200, 0}{+0.72} & \textcolor[RGB]{0, 200, 0}{+0.69} & \textcolor[RGB]{0, 200, 0}{+1.34} \\
\bottomrule
\end{tabular}
\label{Performance on gene mutation prediction}
\end{table}

\subsection{Ablation Studies and Discussion}
In this section, we discuss the effect of two ambiguous designs in SlideGCD. \textit{1) Why use distillation instead of fusion strategies? 2) Is the hypergraph representation necessary? Can we use the simple graph instead?}

\begin{table}
\centering
\caption{Ablation on distillation and fusion.}
\begin{tabular}{l|ccc}
\toprule
\multirow{2}*{Interaction Strategy} & \multicolumn{3}{c}{TCGA-BRCA (Subtyping)} \\ 
& ACC(\%) & AUC(\%) & F1(\%) \\ \midrule

DTFDMIL & 89.30±0.44 & 90.08±0.86 & 83.17±1.43 \\ \midrule

\multicolumn{4}{c}{SlideGCD-DTFDMIL} \\ \midrule

w LogitsAddFusion & 89.57±2.79 & 90.57±2.45 & 83.01±5.03 \\

w FeatCatFusion & 87.31±0.71 & 86.80±1.94 & 79.25±1.71 \\

w FeatAddFusion & 88.90±0.88 & 90.17±0.92 & 82.12±1.16 \\

w Distillation (KLDiv) & \textbf{91.89±1.49} & \textbf{93.35±1.31} & \textbf{87.20±2.17} \\

w Distillation (JSDiv) & \underline{91.50±0.50} & \underline{92.83±0.93} & \underline{86.52±0.78} \\
\bottomrule
\end{tabular}
\label{Ablation on distillation and fusion}
\end{table}

\subsubsection{Distillation Vs. Fusion} There are many ways to gather knowledge from multiple branches of a neural network. The first idea is the Fusion strategy including Feature-level Fusion and Logits-level Fusion. In our case, the fusion strategy could be a substitute as long as it is well-performed and does not introduce new parameters. Therefore, comparisons were made between five classic strategies in Table \ref{Ablation on distillation and fusion} including both fusion and distillation: \textit{i) \textbf{LogitsAddFusion}}: The outputted logits from both branches are added together for final predictions. \textit{ii) \textbf{FeatCatFusion}}: The embeddings outputted from the MIL backbone and the final graph convolutional layer are concatenated and then inputted to a linear layer to generate final logits for predictions. \textit{iii) \textbf{FeatAddFusion}}: The same embeddings are sent to a project layer to align channels, and then added together to input a linear layer to generate final logits for predictions. \textit{iv) \textbf{Distillation (KLDiv)}}: The $L_{JS}$ is replaced by the KL divergence loss following \cite{hinton2015distilling}. \textit{v) \textbf{Distillation (JSDiv)}}: The full version of SlideGCD.
As shown in Table \ref{Ablation on distillation and fusion}, only the LogitsAddFusion strategy slightly improved DTFDMIL with 89.57\% ACC and 90.57\% AUC among these three fusion strategies. Yet both common distillation strategies made significant performance rises for DTFDMIL without bringing extra computations. It demonstrates that the distillation is more suitable on this occasion that the two branches sensibly concern different aspects of information. Besides, the additional symmetry of JS divergence compared to KL divergence makes the network more stable and robust, which is reflected in the minor standard deviations at approximately equal metrics.

\subsubsection{Hypergraph Vs. Simple Graph} The difference between a hypergraph with a simple graph is their definition of edge where the hyperedge can connect more than two nodes. From this perspective, the simple graph can be viewed as a special case of hypergraph. Then could the SlideGCD pipeline take effect without the hypergraph representation and the hypergraph convolutional layers? To answer this question, we conducted comparisons that replaced the hypergraph and related layers with the simple graph and three widely used graph convolution layers. As shown in Table \ref{Ablation on different graph convolution operation}, it is gratifying that all modifications of the SlideGCD work properly and bring visible performance increase to the baseline. In conclusion, we have demonstrated the improvements brought by SlideGCD are not bound to any specific graph convolution operation but are achieved by the framework itself.

\begin{table}
\centering
\caption{Ablation on different graph convolution operation.}
\begin{tabular}{l|ccc}
\toprule
\multirow{2}*{GraphConv Operation} & \multicolumn{3}{c}{TCGA-BRCA (Subtyping)} \\ 
& ACC(\%) & AUC(\%) & F1(\%) \\ \midrule

DTFDMIL & 89.30±0.44 & 90.08±0.86 & 83.17±1.43 \\ \midrule

\multicolumn{4}{c}{SlideGCD-DTFDMIL} \\ \midrule

w GCNConv \cite{kipf2016semi}
& \underline{90.76±0.74} & 91.27±1.25 & \underline{84.74±1.13} \\

w GATConv \cite{velickovic2017graph}
& 90.17±0.77 & \underline{91.57±1.94} & 83.74±1.90 \\

w GINConv \cite{xu2018powerful}
& 89.83±2.67 & 91.03±2.13 & 84.42±3.75 \\

w HyperGraphConv \cite{bai2021hypergraph}
& \textbf{91.50±0.50} & \textbf{92.83±0.93} & \textbf{86.52±0.78} \\
\bottomrule
\end{tabular}
\label{Ablation on different graph convolution operation}
\end{table}

\begin{figure}
\centerline{\includegraphics[width=1\columnwidth]{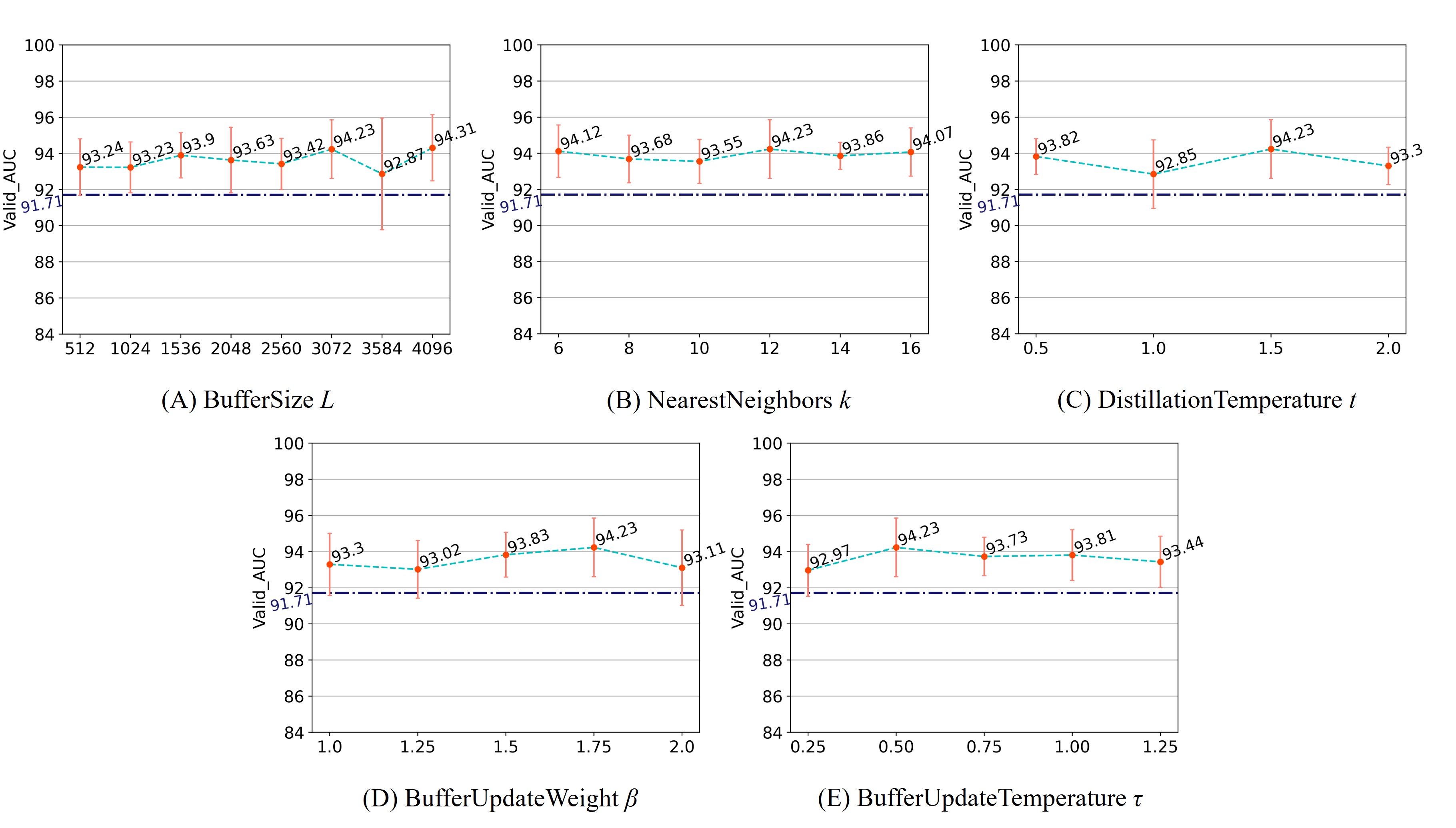}}
\caption{Performance curves on the validation dataset in the five-fold cross-validation, where the error bar indicates the standard deviation of the metrics and the dashed line parallel to the horizontal axis represents the metrics that the baseline (DTFDMIL) can achieve.}
\label{Hyper-parameters}
\end{figure}

\begin{figure}
\centerline{\includegraphics[width=1\columnwidth]{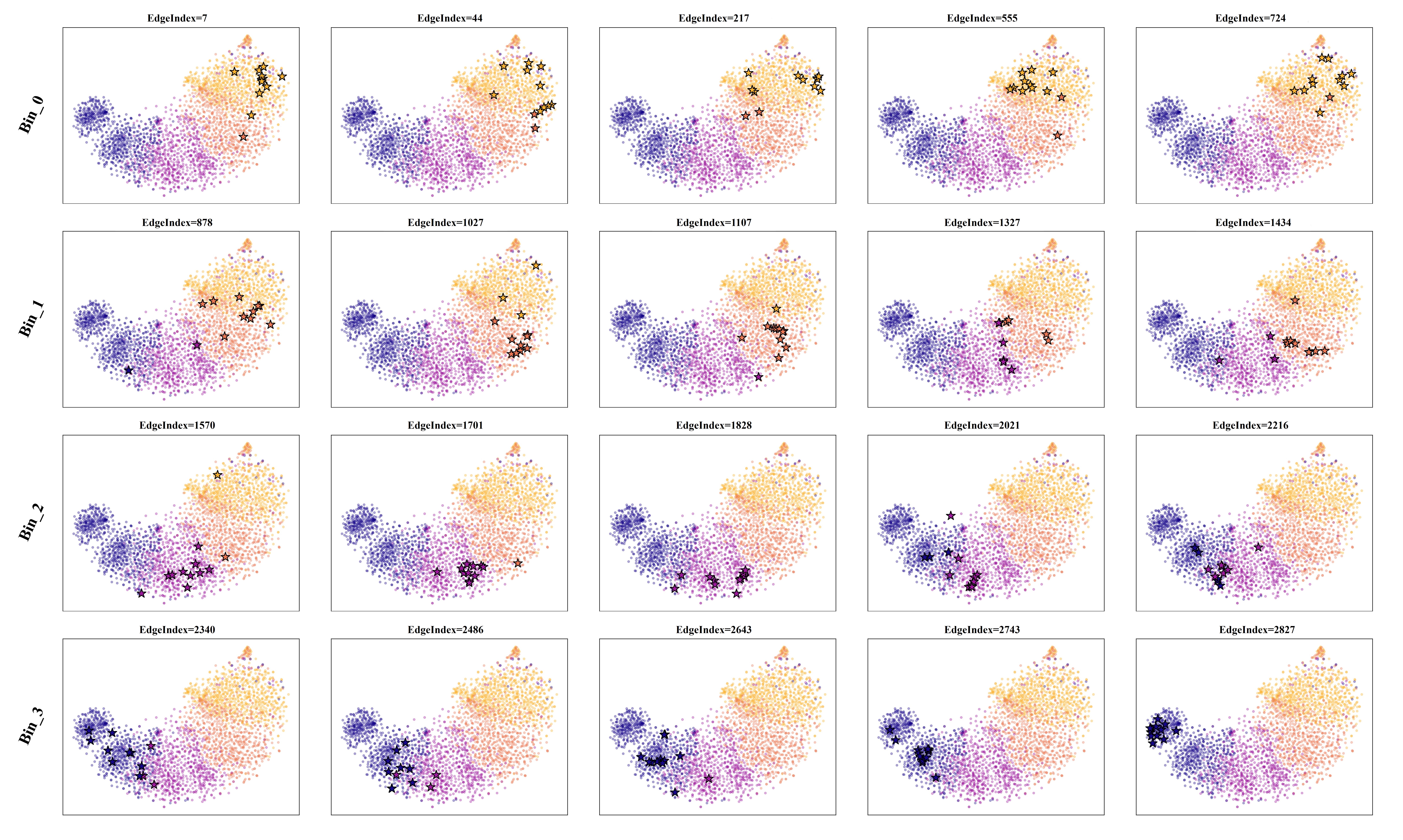}}
\caption{T-SNE visualizations of the overall distribution of the nodes from the node buffer and 20 highlighted hyperedges for TCGA-BLCA. Each point corresponds to a patient and its color goes deeper as the survival time increases. The stars in each sub-figure make up the hyperedge.}
\label{Visualization}
\end{figure}

\subsection{Hyper-parameters Verification}
In this section, we systematically explore the effect of hyper-parameters in SlideGCD with the backbone of DTFDMIL on the TCGA-BRCA (Subtyping) task. The results come from the five-fold cross-validation on the validation set. The impact of the following hyper-parameters will be discussed: 1) the size of node buffer $L$, 2) the nearest neighbors $k$, 3) the distillation temperature $t$, 4) the loss weight of buffer update $\beta$, and 5) the temperature in buffer update $\tau$.
 
\subsubsection{Size of node buffer $L$} $L$ determines the exact number of nodes in the slide-based graph and greatly affects the capacity of the class-aware sub-queue. When $L$ is set too large, the node buffer will contain much more outdated slide embeddings that might defect the performance of the SlideGCD and will introduce more extra computation as well. When it is too small, the effect of the slide-based graph might be inapparent since there is a lack of sufficient homogeneous information in the slide-based graph. We tuned $L$ in the range of [1024, 4096] with a step of 512. The curve in Fig. \ref{Hyper-parameters}(A) shows that the performance of SlideGCD is indeed altered, but it can still stably surpass the baseline. Eventually, we choose $L=3072$ as the optimal value for balancing the accuracy and computational complexity.

\subsubsection{Nearest neighbors $k$} $k$ is another important hyper-parameter that controls the connection density in the slide-based graph and thus influences the message passing during graph learning. Experimentally, we tuned $k \in [6, 8, 10, 12, 14, 16]$ for verification. From Fig. \ref{Hyper-parameters}(B), it is known that SlideGCD is less sensitive to $k$ than $L$ since the validated AUCs are quite steady and always at a relatively high level compared with the baseline. We choose $k=12$ as the default setting for a better AUC.

\subsubsection{Distillation temperature $t$} This hyper-parameter manages the effect of distillation. In a higher temperature situation (i.e. $t>1$), distillation focuses on transferring knowledge from the teacher model. When getting a lower temperature (i.e. $t<1$), distillation tends to alleviate the impact of noise in negative samples \cite{hinton2015distilling}. In our application, our objective is to transfer the well-learned knowledge in MIL head to the SlideGNN, thus a relatively large temperature coefficient should have a better effect. Following the results in Fig. \ref{Hyper-parameters}(C), we set $t=1.5$.

\subsubsection{Loss weight of buffer update $\beta$} $\beta$ controls the contribution from the buffer update. The balance between each component of the final loss is important for achieving the best performance. We tested it in the range of [1.0, 2.0] with a step of 0.25 since the buffer update is not the main supervision signal in our framework. As shown in Fig. \ref{Hyper-parameters}(D), the model performance is not sensitive to it. Empirically, we set $\beta=1.75$.

\subsubsection{Temperature in buffer update loss $\tau$} $\tau$ supervises the effect of contrastive learning. In SlideGCD, the node embeddings continuously shift during training resulting in many noises and ultimately leading to heterogeneity in the slide-based graph. Under the circumstances, we wish the buffer update strategy which is based on contrastive learning could alleviate the impact of noise brought by the over-shifting slide embeddings. The curves in Fig. \ref{Hyper-parameters}(E) validate the rationality of the motivation since the performances reach the peak with a lower temperature $\tau=0.5$.

\subsection{Interpretability and Visualization}
To intuitively discuss the effect of the slide-based graph learning, we visualize the overall distribution of the node embeddings within the node buffer (on TCGA-BLCA) via T-SNE where we highlight 20 hyperedges to assess its node distribution. Following the discretization of survival time in \cite{chen2021whole}, the patients are normally divided into 4 buckets (bins: 0-3) as the increment of survival time. As shown in Fig. \ref{Visualization}, all sub-figures share the same background which represents the overall distribution of the node embeddings in the buffer where each point indicates a node, and the color goes deeper as the survival time extends. The pointed stars are from the same hyperedge and the index is above each subgraph.

From the overall background of Fig. \ref{Visualization}, it can be clearly observed that the class-aware node buffer can significantly separate the nodes from different buckets and the distribution of the node embeddings follows a pattern that the survival time of nodes gradually extends in a circular arc shape from top right to bottom and then to left. Inspecting each sub-figure, nodes are more inclined to associate with other nodes in the same bucket (having close survival times). As the survival time increases (the edge index grows), the center of the hyperedge is shifting along that pattern. The above observations intuitively validate that the proposed pipeline has achieved the association of homogeneous samples and also prove the improvement brought by SlideGCD is explainable and traceable.

\section{Conclusion}
\label{Conclusion}
In this paper, we present an end-to-end generic pipeline SlideGCD for histopathology whole slide image analysis, which exploringly takes the unrestricted slide inter-correlations into account via the slide-based graph and proves its consistent improvements. The rehearsal-based adaptive graph construction strategy we devised, models the WSI dataset into a slide-based graph where WSIs are nodes and their connections can be updated during training. Besides, knowledge distillation is applied to train MIL and the graph branch collaboratively and try not to lose the inner-contextual knowledge learned by Head$_{MIL}$ as much as possible. Extensive experiments and visualizations are conducted to demonstrate the effectiveness and robustness in an interpretable way.

Although the proposed SlideGCD achieved promising improvements on various backbones and downstream tasks, some upgrades still could be made. For example, buffer update strategies based on other principles, such as uncertainty, may optimize the training and inference overhead. The proposed pipeline can significantly benefit from the development of the rehearsal buffer. We hope that this work can attract much attention to the exploration of the slide-level interaction and we believe this will assist in the advancement of foundation models for computational pathology.

\end{document}